\documentclass[10pt,twocolumn,letterpaper]{article}

\usepackage[pagenumbers]{cvpr} 

\definecolor{cvprblue}{rgb}{0.21,0.49,0.74}
\usepackage[pagebackref,breaklinks,colorlinks,allcolors=cvprblue]{hyperref}
\usepackage{hyperref}       
\usepackage{url}            
\usepackage{booktabs}       
\usepackage{amsfonts}       
\usepackage{nicefrac}       
\usepackage{microtype}      
\usepackage{xcolor}         
\usepackage{amsmath, amssymb}
\usepackage{array} 
\usepackage{wrapfig}
\usepackage{multirow}

\usepackage{graphicx}      
\usepackage{subcaption}    
\usepackage{caption}       
\usepackage{float}         
\usepackage[table]{xcolor}

\usepackage{enumitem}  
\usepackage{subcaption}  
\usepackage{algorithmic}
\usepackage{algorithm}
\usepackage{multirow}
\usepackage{diagbox}

\def\onedot{.\xspace}

\usepackage{xspace}

\def\onedot{.\xspace}
\def\eg{\emph{e.g}\onedot} 
\def\ie{\emph{i.e}\onedot}

\def\paperID{11259} 
\def\confName{CVPR}
\def\confYear{2026}

\title{TAS-LoRA: Transformer Architecture Search with Mixture-of-LoRA Experts}

\author{Jeimin Jeon$^{1,2}$ \quad\quad\quad Hyunju Lee$^{1}$ \quad\quad\quad Bumsub Ham$^{1,3}$\thanks{Corresponding author.} \vspace*{3mm}\\
$^{1}$Yonsei University \quad $^{2}$Articron Inc. \quad $^{3}$Korea Institute of Science and Technology (KIST) \\
{ \url{https://cvlab.yonsei.ac.kr/projects/TAS-LoRA/}}
}

\begin{document}
\maketitle
\begin{abstract}
    Transformer architecture search~(TAS) discovers optimal vision transformer~(ViT) architectures automatically, reducing human effort to manually design ViTs. However, existing TAS methods suffer from the feature collapse problem, where subnets within a supernet fail to learn subnet-specific features, mainly due to the shared weights in a supernet, limiting the performance of individual subnets. To address this, we propose TAS-LoRA, a novel method that introduces parameter-efficient low-rank adaptation~(LoRA) to enable subnet-specific feature learning, while maintaining computational efficiency. TAS-LoRA incorporates a Mixture-of-LoRA-Experts~(MoLE) strategy, where a lightweight router dynamically assigns LoRA experts based on subnet architectures, and introduces a group-wise router initialization technique to encourage diverse feature learning across experts early in training. Extensive experiments on ImageNet and several transfer learning benchmarks, including CIFAR-10/100, Flowers, CARS, and INAT-19, demonstrate that TAS-LoRA mitigates feature collapse effectively, improving performance over state-of-the-art TAS methods significantly.
    \end{abstract}

    \section{Introduction}

    Transformer architectures~\cite{dosovitskiy2020image,touvron2021training,liu2021swin,vaswani2017attention} have demonstrated remarkable success beyond natural language processing, particularly in computer vision, surpassing convolutional neural networks (CNNs)~\cite{he2016deep,tan2019efficientnet,howard2017mobilenets,sandler2018mobilenetv2} in various tasks. Unlike CNNs, Vision transformers~(ViTs)~\cite{dosovitskiy2020image,touvron2021training,liu2021swin} leverage self-attention mechanisms to capture long-range dependencies and global context, allowing them to model complex spatial relationships more effectively. Despite these advantages, designing optimal ViT architectures for different hardware constraints remains challenging, as it requires extensive manual tuning, making deployment computationally expensive and limiting scalability.

    To address this, transformer architecture search (TAS) has emerged as a promising solution, automating the discovery of ViT architectures optimized for specific constraints, reducing the human effort significantly. Among TAS approaches~\cite{lee2024assembling,liu2022focusformer,chen2021autoformer,wei2024auto,su2022vitas}, one-shot TAS methods~\cite{chen2021autoformer,liu2022focusformer,su2022vitas} have gained particular attention due to their efficiency. They train a weight-entangled supernet, where all possible subnets share a common set of weights, and leverage it to estimate the performance of different subnet architectures. The selected subnet can directly be deployed without additional training, significantly reducing computational costs. Despite their efficiency, weight-entangled supernets suffer from the feature collapse problem, where subnets within supernets fail to learn distinct feature representations~(Fig.~\ref{fig:teaser}(a-b)). Since the supernet is optimized to minimize an overall loss across all possible subnets, gradients from different subnets are aggregated to update the shared weights in the supernet. Accordingly, the shared weights enable learning generic features that work well for diverse subnets~\cite{deng2020understanding}, but this prevents individual subnets from fully leveraging their architectural characteristics, leading to suboptimal performance.

    

    To overcome the feature collapse problem, we propose a novel TAS approach to learning feature representations tailored to each subnet, dubbed TAS-LoRA, using a parameter-efficient low-rank adaptation (LoRA) technique. Unlike conventional methods that rely on shared weights across all subnets, TAS-LoRA introduces trainable low-rank parameters to each subnet, allowing subnets to learn subnet-specific feature representations~(Fig.~\ref{fig:teaser}(c)). Additionally, at inference, the LoRA parameters can be merged into each subnet, introducing no additional computational overhead. However, directly assigning a unique LoRA to every subnet would be computationally prohibitive due to the enormous number of possible subnet configurations. To address this, we propose to incorporate Mixture-of-LoRA-Experts (MoLE), which efficiently distributes a limited set of LoRA experts across subnets. Instead of assigning a unique LoRA to each subnet, our approach employs a router that dynamically selects and combines relevant LoRA experts based on the subnet architecture for learning subnet-specific feature representations. One major challenge is that, during the early stages of training, the router initially treats all LoRA experts equally, resulting in similar feature representations and redundancy among experts. To address this issue, we use a group-wise router initialization strategy, where transformer blocks within subnets are grouped based on architectural similarities, and the router is initialized such that each LoRA expert is assigned to distinct groups. We validate our method through extensive experiments on ImageNet~\cite{deng2009imagenet} and evaluate its transferability across multiple datasets, including CIFAR-10/100~\cite{krizhevsky2009learning}, Flowers~\cite{nilsback2008flowers}, CARS~\cite{krause2013cars}, and INAT-19~\cite{van2018inaturalist}. Our results demonstrate that TAS-LoRA significantly outperforms state-of-the-art approaches, confirming its effectiveness in mitigating the feature collapse problem and improving the performance of TAS. We summarize our contributions as follows:
    \begin{itemize}[leftmargin=*]
        \item We propose TAS-LoRA, a novel approach that mitigates the feature collapse problem in TAS using LoRA, while learning subnet-specific features.
        \item To enhance efficiency, we propose to incorporate a MoLE for TAS, which dynamically assigns and combines LoRA experts based on each subnet architecture. We also introduce group-wise router initialization to encourage that LoRA experts learn distinct feature representations early in training.
        \item Extensive experiments on various datasets demonstrate the superiority of TAS-LoRA over the state-of-the-art methods.
    \end{itemize}
    
    \section{Related work}
    
    \subsection{Transformer architecture search}
    Early works on neural architecture search~(NAS)~\cite{liu2019darts,bender2018understanding,cai2019proxylessnas,cai2020once,oh2024efficient} focus on handling CNN architectures. With the growing interest in ViTs, research on searching for optimal ViT architectures has gained significant attention. TAS can be categorized into zero-shot and one-shot TAS methods.
    
    Zero-shot TAS approaches~\cite{zhou2022training,lee2024assembling,sun2023unleashing} leverage zero-cost proxies to predict the network performance without training. For example, TF-TAS~\cite{zhou2022training} introduces the DSS-indicator to estimate the characteristics of self-attention and MLP modules. $\mathcal{E}$-GSNR~\cite{sun2023unleashing} exploits the gradient signal-to-noise ratio to predict generalization and convergence abilities of the network. AZ-NAS~\cite{lee2024assembling} combines multi zero-cost proxies to evaluate networks from various perspectives, such as feature expressivity and network complexity. Zero-shot TAS methods are highly efficient for selecting network architectures, but they require to train the chosen architectures from scratch, prior to deployment, which is computationally demanding, especially when adjusting for different hardware constraints.
    
    One-shot TAS approaches~\cite{chen2021autoformer,su2022vitas,liu2022focusformer,PreNAS,jeon2025subnet, jawahar2024mos} instead train a supernet encompassing all possible subnets, deploying the subnets without additional training. Specifically, AutoFormer~\cite{chen2021autoformer} optimizes the supernet by randomly sampling subnets and updating their parameters. VITAS~\cite{su2022vitas} shows that AutoFormer results in imbalanced updates across channels, leading to suboptimal performance. To address this issue, it proposes a new supernet construction strategy to encourage uniform updates across all channels. FocusFormer~\cite{liu2022focusformer} replaces random sampling with learnable subnet samplers, which jointly optimize the sampling process and supernet weights. DYNAS~\cite{jeon2025subnet} proposes a dynamic training strategy that adjusts learning rates and momentum buffers based on the sampled subnet. A major limitation of all aforementioned methods is that significant numbers of subnets share the same weights, leading to the feature collapse problem, where subnets fail to learn specialized features.
    
    For effective supernet training, PreNAS~\cite{PreNAS} proposes to reduce the number of subnets considered in a supernet. Specifically, it exploits zero-cost proxies to pre-select a very few subnets for constructing the supernet,~\eg, 6 out of $2\times10^{8}$ possible architectures in the Autoformer-T space. However, similar to zero-shot TAS approaches, it struggles to adapt to new hardware constraints, such as FLOPs limit, if they are not accounted for the selected subnets. When new constraints are given, a different set of subnets should be chosen, and the supernet needs to be retrained from scratch. In contrast, our method constructs a supernet containing all possible subnets, similar to conventional one-shot TAS approaches, while allowing  each subnet to learn specialized features using LoRA. To the best of our knowledge, this is the first work to exploit MoLE for TAS.

    Most related to ours, MoS~\cite{jawahar2024mos} introduces a mixture-of-supernets strategy to reduce the gradient conflict among subnets, similar to few-shot NAS~\cite{oh2024efficient,zhao2021few}. Our method differs from MoS in the following aspects: (1) while MoS aims to minimize gradient conflicts, our primary goal is to address the feature collapse problem in TAS; (2) MoS requires training multiple supernets, resulting in high computational costs, whereas our approach trains a lightweight set of MoLE layers efficiently; (3) MoS trains all supernets from scratch, while ours fine-tunes the MoLE layers on a pretrained supernet; and (4) unlike MoS, we introduce an LSTM-based router to capture sequential dependencies across transformer blocks, and apply group-wise router initialization to learn diverse feature representations.
    
    \subsection{LoRA} 
    LoRA~\cite{hu2022lora} is a technique that fine-tunes pretrained models efficiently with low-rank learnable matrices, while keeping the original weights frozen. This significantly reduces the number of trainable parameters, making fine-tuning computationally efficient. Several variants of LoRA have been proposed to enhance its effectiveness. DoRA~\cite{liu2024dora} decouples pretrained weights into two separate learnable parameters, magnitude and direction, to enhance the learning capacity. PISSA~\cite{meng2025pissa} initializes the LoRA branch using the principal components of the pretrained weights for faster convergence. AdaLoRA~\cite{zhang2023adalora} adjusts the rank of layers based on their importance,~\ie,~the layers with higher importance are assigned higher ranks, and vice versa, for better performance. 
    
    More recently, MoLE~\cite{wu2024mixture,Buehler_XLoRA_2024,yang2024multi} has been proposed, where a router dynamically selects LoRA experts based on input feature maps. While effective, the router in the existing MoLE-based methods should recompute expert weights for every input, leading to significant computational overhead. Moreover, selected experts vary w.r.t the input, and thus LoRA parameters could not be integrated into pretrained weights. Our approach addresses this inefficiency by exploiting subnet configurations as router inputs, instead of feature maps. Since the selected subnet remains fixed, the router runs only once, independent to the input. This also allows LoRA parameters to be merged into the supernet, further reducing inference overhead.
    
    \subsection{Feature collapse problem}
    Feature collapse refers to a phenomenon that different architectures with shared weights are likely to provide similar feature representations, limiting the ability to exploit their architectural characteristics. UENSA~\cite{deng2020understanding} has shown the feature collapse problem with stochastic networks, where network architectures change randomly during training and inference~(\eg, varying skip connection paths between layers), while they share the same set of weights. To alleviate this, it introduces architecture-specific parameters enforcing slight variations in feature learning, which is however tailored to CNNs, and considers five candidates of network architectures only. As the number of candidate architectures increases, the number of architecture-specific parameters grows linearly, requiring additional memory resources. The total number of training iterations also increase to optimize parameters of each candidate sufficiently, resulting in higher computational costs. To our knowledge, we are the first to identify and address the feature collapse problem in TAS. Our method uses a small set of LoRA experts, combined with a router, that selects appropriate experts dynamically based on the characteristics of subnets. This enables learning specialized features for individual subnets efficiently, even for supernets with an enormous number of subnets.

    A related but contrasting observation comes from Pi-NAS~\cite{peng2021pi} that subnets within its supernet show diverse feature representations. This discrepancy arises from the difference in supernet construction. Pi-NAS employs a weight-sharing strategy, where different operations (\eg, Conv $3\times3$ and Conv $1\times1$) within the same layer use separate branches with completely independent parameters, which may lead to distinct feature representations for each branch. Note that this strategy requires retraining after subnet selection. Recently, weight-entangled strategies are more popular in TAS~\cite{chen2021autoformer,su2022vitas,PreNAS,liu2022focusformer}, due to the efficiency in eliminating the retraining process. In weight-entangled strategies, the weights of operations are sampled from a shared parameter set per layer. During training, different subnets update the shared weights of each layer, preventing the parameters from specializing for each architecture. This limits learning subnet-specific features, and leads to the feature collapse problem.

\section{Method}
In this section, we briefly describe the problem of TAS~(Sec.~\ref{sec:problem}), present our empirical analysis of the feature collapse problem in weight-entangled supernets~(Sec.~\ref{sec:empirical_analysis}), and then introduce our TAS-LoRA method~(Sec.~\ref{sec:tas-lora}).

\subsection{Preliminaries}
\label{sec:problem}

In TAS, a supernet $S$ is an over-parameterized model that encompasses a large set of possible subnets. Each subnet is characterized by two types of attributes: (1) block-level attributes, which vary across transformer blocks~(\eg,~the number of attention heads, MLP ratio), and (2) subnet-level attributes, which remain consistent across all blocks~(\eg,~an embedding dimension, subnet depth). Formally, we denote the set of all subnets within the supernet as $\mathcal{N}$, where each subnet $\mathcal{N}_i$ is defined as:
\begin{equation} 
\mathcal{N}_i = (v, e, A), 
\end{equation}
where \( v \) represents the depth of the subnet,~\ie,~the number of transformer blocks, and \( e \) denotes the embedding dimension. Note that these subnet-level attributes are shared across all blocks. \( A \) captures the block-specific attributes, defined as
\begin{equation} 
A = \{ (n^{(b)}, m^{(b)}) \}_{b=1}^{v}, 
\end{equation}
where \( n^{(b)} \) and \( m^{(b)} \) represent the number of attention heads and MLP ratio in the $b^{th}$ block, respectively.

During training, a single subnet $\mathcal{N}_i$ is sampled per iteration and optimized using a standard loss function:
\begin{equation} \theta^* = \arg\min_\theta \mathbb{E}_{\mathcal{N}_i \sim \mathcal{N}} [\mathcal{L}(\mathcal{N}_i; \theta)], \end{equation} 
where $\theta$ represents the shared weights in the supernet, and $\mathcal{L}$ denotes a loss function~(\eg,~the cross-entropy loss).

\begin{figure*}[t]
    \centering
    \captionsetup{font={small}}
    \hfill
    \begin{subfigure}[t]{0.2\linewidth}
        \centering
        \includegraphics[width=\textwidth]{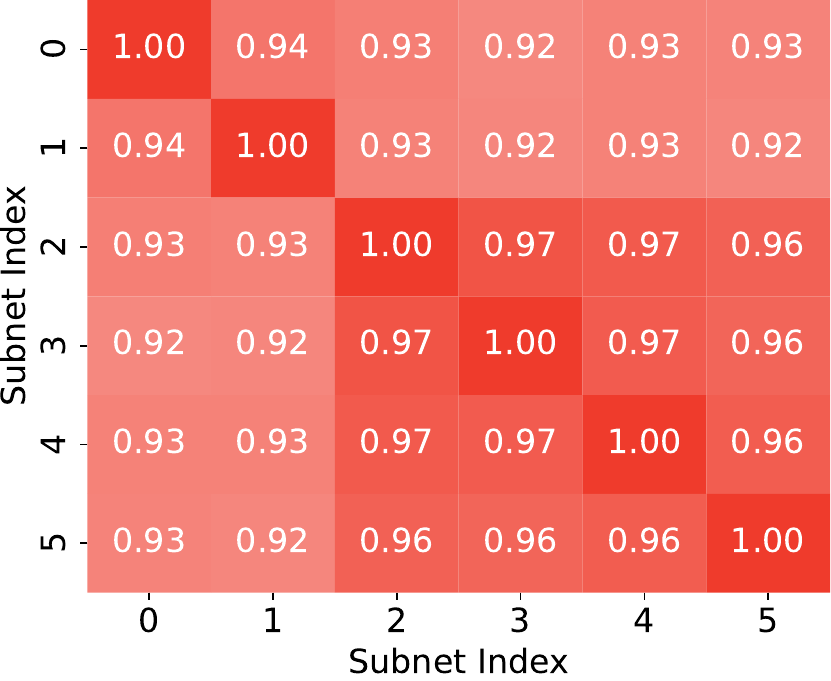}
        \vspace{-0.5cm}
        \caption{AutoFormer~\cite{chen2021autoformer}}
        \label{fig:heatmap_auto}
    \end{subfigure}
    \hfill
    \begin{subfigure}[t]{0.2\linewidth}
        \centering
        \includegraphics[width=\textwidth]{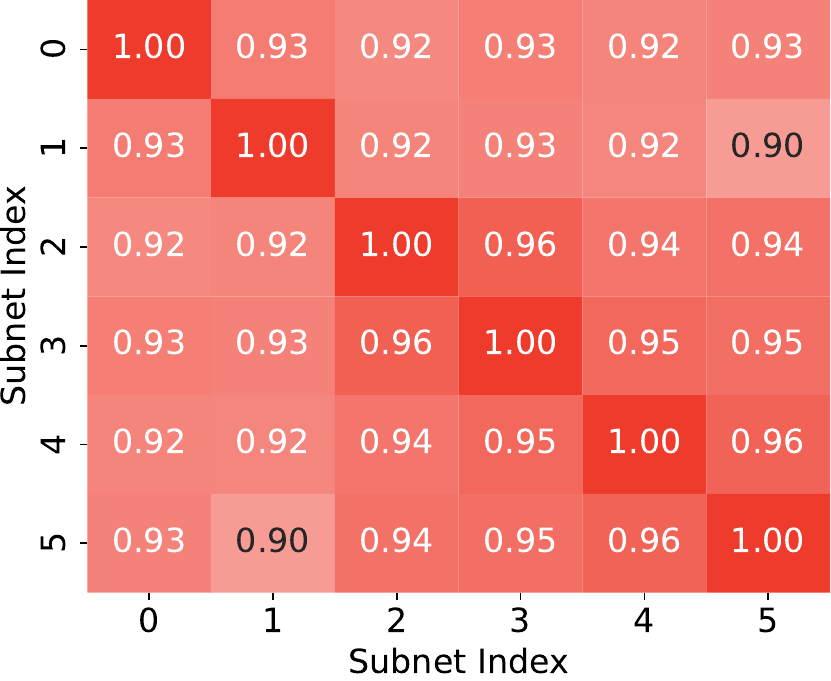}
        \vspace{-0.5cm}
        \caption{DYNAS~\cite{jeon2025subnet}}
        \label{fig:heatmap_dynas}
    \end{subfigure}
    \hfill
    \begin{subfigure}[t]{0.2\linewidth}
        \centering
        \includegraphics[width=\textwidth]{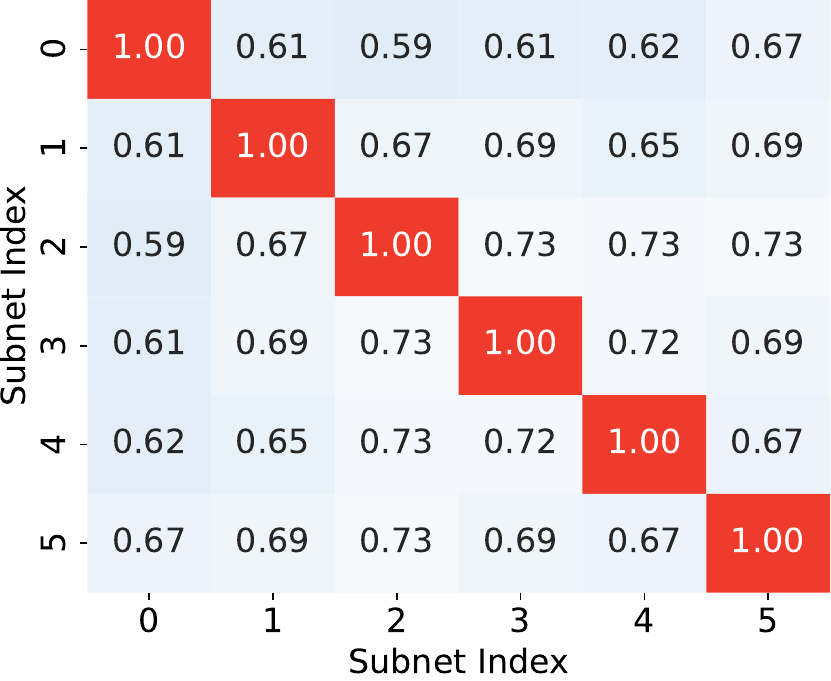}
        \vspace{-0.5cm}
        \caption{TAS-LoRA (Ours)}
        \label{fig:heatmap_ours}
    \end{subfigure}
    \hfill
    \begin{subfigure}[t]{0.2\linewidth}
        \centering
        \includegraphics[width=\textwidth]{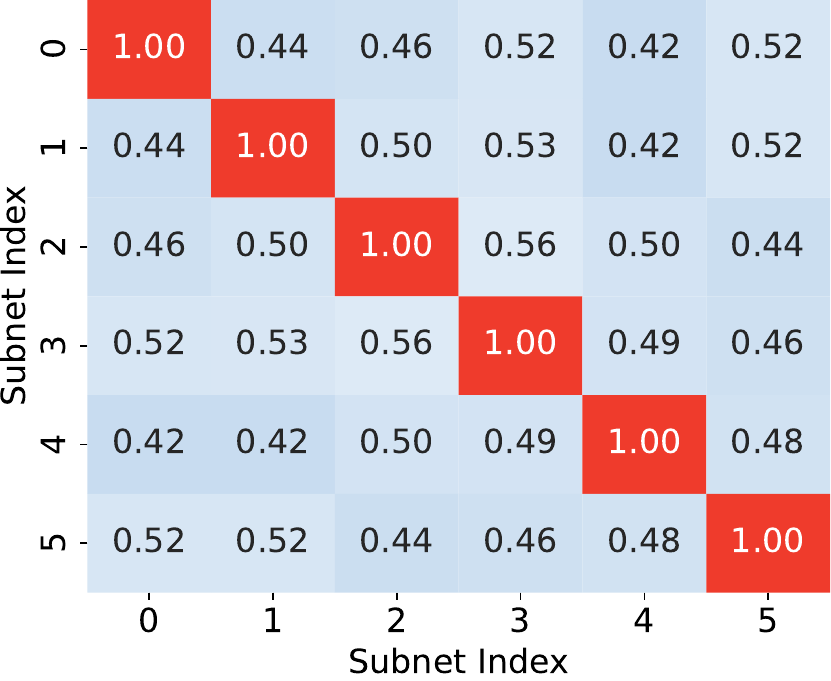}
        \vspace{-0.5cm}
        \caption{Independent}
        \label{fig:heatmap_scratch}
    \end{subfigure}
    \hfill
    \begin{subfigure}[t]{0.02\linewidth}
        \centering
        \vspace{-2.2cm} \includegraphics[width=\textwidth]{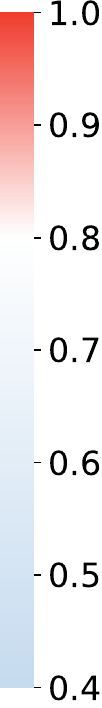}
    \end{subfigure}
    \vspace{-0.2cm}

    \caption{
    Feature similarities between subnets trained with different strategies. Six subnets are randomly sampled from each supernet, and average cosine similarity is computed between features from the penultimate-layer:~(a)~AutoFormer~\cite{chen2021autoformer}, (b) DYNAS~\cite{jeon2025subnet}, (c) TAS-LoRA, (d) Subnets trained independently from scratch.
        }

    \vspace{-0.4cm}

    \label{fig:teaser}
\end{figure*}

\vspace{-0.1cm}
\subsection{Empirical analysis}
\label{sec:empirical_analysis}
\vspace{-0.1cm}

TAS methods typically adopt a weight-entangled supernet for training, where all subnets share a common set of parameters. While the weight-entangled supernet reduces training cost significantly, it suffers from the feature collapse problem that different subnets within the supernet tend to provide highly similar feature representations, regardless of their architectural differences. Since the supernet optimizes for an overall loss across all subnet configurations, the shared weights attempt to capture generic features that work across subnets, rather than the features tailored to each subnet, resulting in suboptimal performance for individual subnets.

We visualize in Fig.~\ref{fig:teaser} (a-b) cosine similarity between feature representations from six subnets, randomly sampled and trained with weight-entangled TAS approaches, AutoFormer~\cite{chen2021autoformer} and DYNAS~\cite{jeon2025subnet}. We can see that feature representations are nearly identical. Subnets trained independently from scratch, however, exhibit significantly lower similarity, suggesting that optimal feature representations vary, according to subnet architectures~(Fig.~\ref{fig:teaser}(d)). This observation is further supported by the top-1 accuracy in Table~\ref{tab:teaser}, where subnets trained via AutoFormer and DYNAS perform notably worse than their counterparts trained from scratch independently.

To address this limitation, we propose TAS-LoRA, which encourages subnets to learn more specialized and diverse features. Rather than relying on shared weights solely, TAS-LoRA introduces lightweight, subnet-specific adaptation modules that reduce feature redundancy across subnets~(Fig.~\ref{fig:teaser}(c)), thereby achieving top-1 accuracies comparable to subnets trained independently from scratch~(Table~\ref{tab:teaser}).

\begin{table}[t]
    \captionsetup{font={small}}
    \caption{Top-1 accuracies (\%) of six subnets, randomly sampled and trained with different strategies. "AutoFormer" and "DYNAS" represent weight-entangled supernets, where all subnets share weights. "TAS-LoRA" incorporates subnet-specific LoRA modules, while "Indep." trains each subnet independently from scratch.}
    \vspace{-0.6cm}
    \label{tab:teaser}
    \small
    \setlength{\tabcolsep}{0.4em}
    \begin{center}
    \resizebox{0.9\linewidth}{!}{%
    \begin{tabular}{ccccc}
    \toprule
    Subnet & AutoFormer~\cite{chen2021autoformer} & DYNAS~\cite{jeon2025subnet} & TAS-LoRA & Indep. \\
    \midrule
    0 & 76.1 & 76.3 & 76.9 & 76.9 \\
    1 & 74.9 & 74.9 & 75.4 & 75.5 \\
    2 & 76.6 & 76.7 & 77.4 & 77.5 \\
    3 & 74.9 & 75.1 & 76.1 & 76.3 \\
    4 & 76.2 & 76.3 & 77.1 & 77.4 \\
    5 & 76.7 & 76.9 & 77.9 & 78.1 \\
    \bottomrule
    \end{tabular}
    }
    \end{center}
    \vspace{-0.55cm}
  \end{table}

\subsection{TAS-LoRA}
\label{sec:tas-lora}
\begin{figure*}[t]
    \captionsetup{font={small}}
    \centering
    \includegraphics[width=0.8\linewidth]{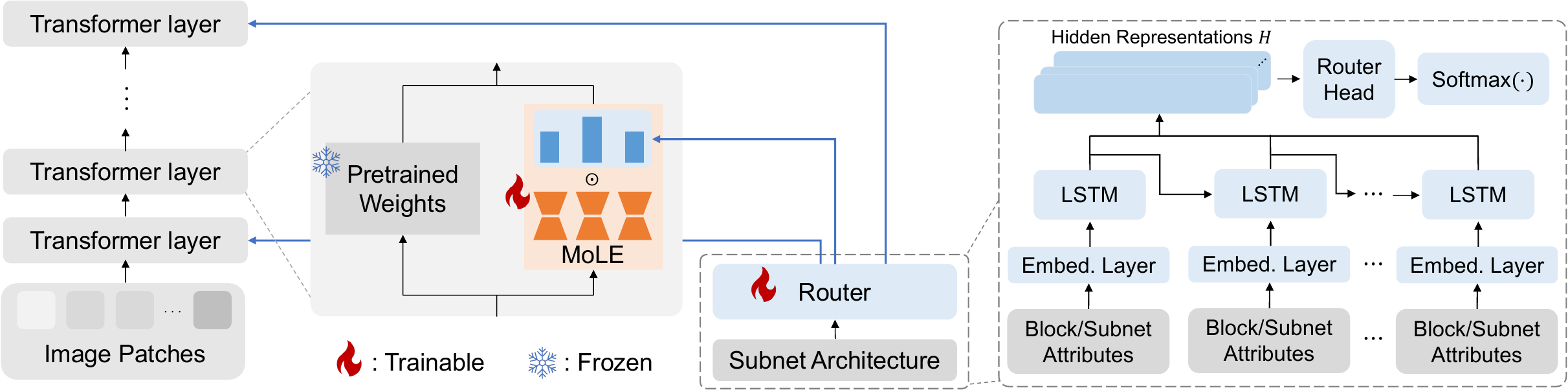}
    \caption{\textbf{Left:} An overview of TAS-LoRA. We exploit a MoLE for TAS to learn subnet-specific feature representations effectively and efficiently. The router dynamically assigns expert weights to each subnet based on its architectural properties. \textbf{Right:} Illustration of our router design. Both block-level and subnet-level attributes are processed by a learnable block embedding layer, and passed through an LSTM~\cite{hochreiter1997long}, which captures sequential dependencies across blocks. The LSTM outputs hidden representations that are further processed by a fully connected router head to generate distributions of expert weights. For more details, please refer to the text.} 
    \vspace{-0.5cm}
    \label{fig:overview}
\end{figure*}

TAS-LoRA exploits LoRA to learn subnet-specific feature representations~(Fig.~\ref{fig:overview}). We conjecture that while a supernet captures shared features across subnets, subnet-specific LoRA enables each subnet to refine these features based on its architectural properties. Assigning a unique LoRA to each subnet however would be computationally expensive, due to the huge number of possible subnet configurations. To mitigate this, we propose MoLE assigning a limited set of LoRA experts to different subnets efficiently. Instead of assigning a separate LoRA to each subnet, our approach exploits a router that dynamically selects and integrates relevant LoRA experts based on subnet architectures, facilitating learning subnet-specific feature representations.

\subsubsection{MoLE}

TAS-LoRA applies MoLE to the linear layers of the supernet, including QKV projection layers and MLP layers in each transformer block. Given $L$ numbers of linear layers, MoLE is applied independently to each, forming $L$ MoLE layers. Each MoLE layer consists of a set of $K$ LoRA experts, where the $k^{th}$ expert in the $l^{th}$ layer is parameterized as $E_k^l = U_k^l D_k^l/r$, with $U_k^l$ and $D_k^l$ being learnable low-rank parameters of the LoRA expert $E_k^l$, respectively, and $r$ denotes the rank of the LoRA. 

For each layer $l$, the router assigns expert weights over the $K$ experts, denoted as $p^l = (p_1^l, p_2^l, \dots, p_K^l)$, where $p^l \in \mathbb{R}^K$ and $\sum_{k=1}^K p_k^l = 1$. A final output for the $l^{th}$ linear layer is then computed as:
\begin{equation} 
y^l = W^l x^l + \sum_{k=1}^{K} p_k^l E_k^l x^l, 
\end{equation}
where $W^l$ represents the weight of an original linear layer, and $x^l$ denotes an input to the $l^{th}$ layer. 

\subsubsection{Router}
\label{sec:router_design}

TAS-LoRA employs an LSTM-based router to assign expert weights dynamically for each layer~(Fig.~\ref{fig:overview}). Each transformer block consists of four linear layers (two in self-attention, two in MLP), suggesting the total number of layers $L$ is $4B$, where $B$ is the number of transformer blocks in the supernet.

The router takes both block-level and subnet-level attributes as input, and provides distributions of expert weights for each layer. A block embedding layer first encodes these attributes into a continuous representation. The encoded block features are then fed into an LSTM~\cite{hochreiter1997long}, which captures sequential dependencies across blocks. The LSTM outputs a hidden representation $h(b) \in \mathbb{R}^{d}$ for the $b^{th}$ block, where $d$ is a hidden dimension of the LSTM, and $b = 1, \ldots, B$\footnote{Note that if a block is not active in a sampled subnet (\ie,~the $b^{th}$ block when $v < b$), the corresponding computations for that block are skipped.}.

The hidden representations from all blocks are concatenated to form a tensor $H\in\mathbb{R}^{B \times d}$, which is passed through a fully connected router head. The router head is parameterized by a learnable weight matrix $R \in \mathbb{R}^{4 \times B \times d \times K}$ and bias term $c \in \mathbb{R}^{4 \times B \times K}$, which together generate the distributions of expert weights $O\in\mathbb{R}^{4B \times K}$. The router applies a softmax function to compute logits for selecting experts for all layers:
\begin{equation} 
    \label{eq:router_logit}
P = \text{softmax}(O, \text{dim}=-1). 
\end{equation}
The expert weight $p^l$ for the $l^{th}$ linear layer is extracted as:
\begin{equation} 
\label{eq:p_l}
p^l = P[l,:].  
\end{equation}
This dynamic routing mechanism allows each layer to select LoRA experts adaptively based on its architectural properties, enabling learning specialized features efficiently\footnote{In our framework, the MoLE layer is also applied to the classifier. Specifically, the hidden representation of the LSTM for the final block is used to compute the expert weights for the classifier. For clarity and to avoid complexity in notation, we omit a description of the MoLE configuration for the classifier.}.


\noindent \textbf{Group-wise router initialization.} 
We have observed that, in the early stages of training, the router tends to assign similar weights to all LoRA experts, which limits expert specialization and causing redundancy. In the works of MoE~\cite{dai2024deepseekmoe,lepikhin2020gshard,fedus2022switch,shazeer2017outrageously}, this issue is typically addressed by introducing auxiliary losses, that encourage different instances within a batch to select different experts. This approach is however not applicable to TAS-LoRA, as our router processes a single subnet only at a time. We instead propose a group-wise router initialization method using architectural attributes of transformer blocks~(\eg,~the number of attention heads).

\begin{figure}[t]
    \captionsetup{font={small}}
    
    \small
    \begin{center}
    \begin{subfigure}[t]{0.47\linewidth}
        \centering
        \includegraphics[width=\linewidth]{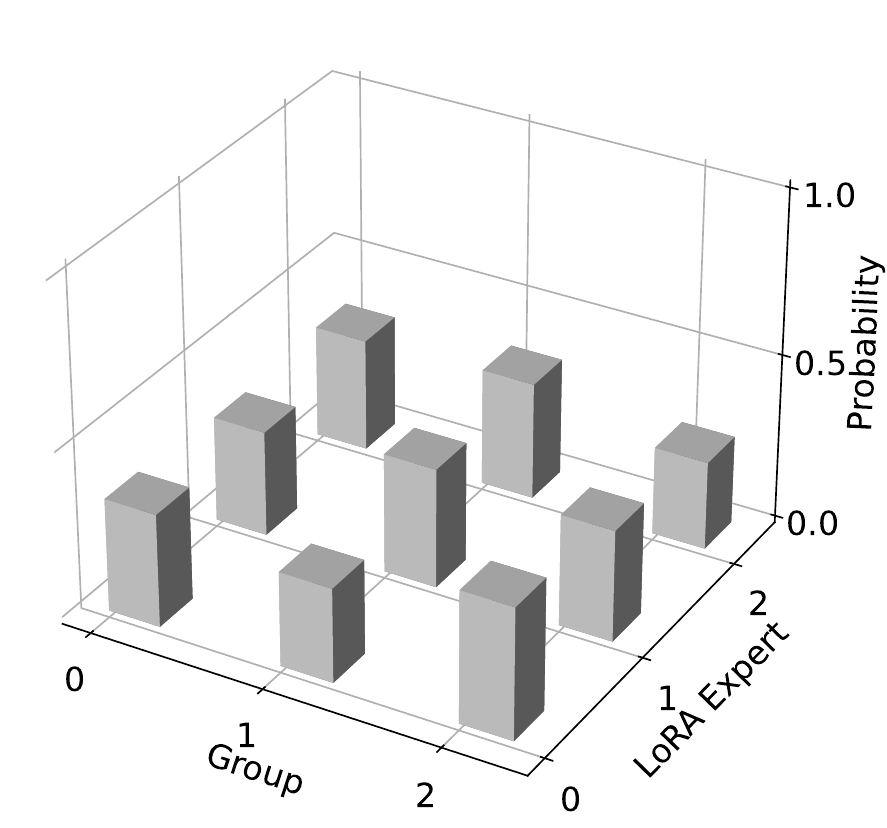}
        \vspace{-0.5cm}
        \caption{Random init.}
    \end{subfigure}
    \begin{subfigure}[t]{0.47\linewidth}
        \centering
        \includegraphics[width=\linewidth]{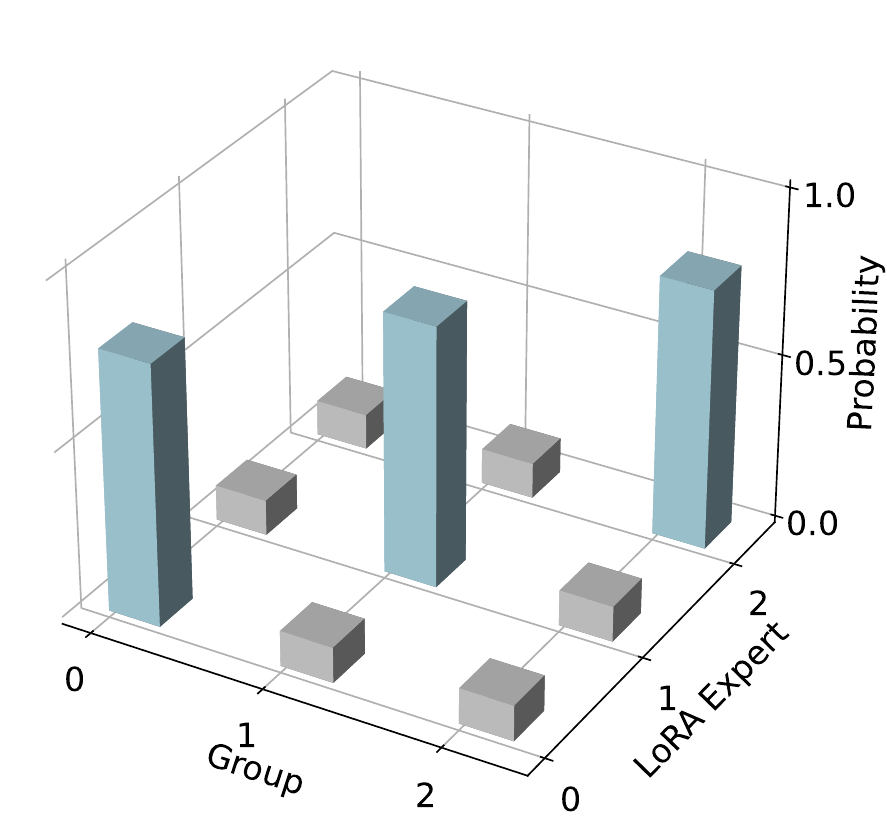}
        \vspace{-0.5cm}
        \caption{Group-wise init.}
    \end{subfigure}
    \end{center}
    \vspace{-0.65cm}
    \caption{Comparison of router initialization strategies. (a)~Random initialization results in nearly uniform assignments of experts, which leads to redundant feature representations. (b)~Group-wise initialization biases selecting experts based on architectural attributes, promoting diverse feature learning across experts.}
    \label{fig:router_init}
    \vspace{-0.2cm}
  \end{figure}

Inspired by the observation that architectures with similar structures tend to learn similar features~\cite{kornblith2019similarity}, we group transformer blocks at the same position across different subnets, with the criteria of the number of attention heads, MLP ratio, and embedding dimension\footnote{We omit the depth attribute, since it does not affect the shape of individual blocks, and exploiting the depth attribute requires more groups, according to the number of depth candidates, without performance improvements.}. That is, these attributes for the criteria are the same within a group. We set the number of LoRA experts to that of groups, and initialize the router, such that each expert is assigned to a distinct group of the blocks. To this end, we use a separate router head for each group~$k$, parameterized by a weight matrix~$R_k' \in \mathbb{R}^{4 \times B \times d \times K}$ and a bias~$c_k' \in \mathbb{R}^{4 \times B \times K}$. We set all elements of $R_k'$ and $c_k'$ to zero, except for a positive value~$\beta$ for the $k^{th}$ element of the bias, where $k$ is an index of the assigned expert for the group~(\ie,~$c_k'[:, :, k] = \beta$). This initialization prevents the hidden representations $H$ from influencing initial expert weights, and thus biases each group toward the designated expert, encouraging exploiting diverse experts early in training ~(Fig.~\ref{fig:router_init}). As training progresses, the router updates expert assignments dynamically, based on the architectural properties of each subnet, allowing the transformer blocks within the same group to select different experts if needed.


\subsubsection{Searching}
\label{sec:searching}


After training the MoLE branch and the router, we search for an optimal subnet within a given search space. The performance of each subnet is evaluated with the shared weights in the supernet and the subnet-specific weights from LoRA. Formally, the optimal subnet \( \mathcal{N}^* \) is obtained by:
\begin{equation}
\mathcal{N}^* = \arg\min_{\mathcal{N}_i \in \mathcal{N}} L_{\text{val}}(\mathcal{N}_i; \theta, \phi),
\end{equation}
where \( L_{\text{val}} \) denotes the validation loss, \( \theta \) represents the supernet\textcolor{red}{} weights, and \( \phi \) includes the parameters of the LoRAs and the router. Note that evaluating all possible subnets is computationally demanding, due to lots of subnet candidates. To efficiently explore the search space, we adopt an evolutionary search strategy~\cite{goldberg1991comparative}, following prior TAS methods~\cite{chen2021autoformer,su2022vitas}. This approach selects and evaluates promising subnets iteratively, identifying optimal architectures efficiently, without testing all configurations exhaustively.

\subsubsection{Inference}
\label{sec:inference}


At inference time, the selected subnet exploits a combination of the shared weights in the supernet and the subnet-specific weights from LoRA. For each linear layer, the merged weight is computed as:
\begin{equation}
W_{\text{merged}}^l = W^l + \sum_{k=1}^{K} p_k^l  E_k^l,
\end{equation}
where \( W^l \) is the original weight from the supernet, and the LoRA experts $E_k^{l}$ are weighted according to their routing probabilities \( p_k^l \). Note that the merged weights are fixed at inference time, suggesting that there are no additional computational costs during inference.

\section{Experiments}
\label{sec:experiments}
In this section, we describe experimental settings, and present results of TAS-LoRA on various datasets~\cite{deng2009imagenet,krizhevsky2009learning,nilsback2008flowers,krause2013cars,van2018inaturalist}. We then provide in-depth analyses of TAS-LoRA. Please refer to the supplement for a detailed analysis of the hyperparameter settings.

\begin{table}[t]
    \captionsetup{font={small}}
    \caption{Configurations of three variants of the AutoFormer space~\cite{chen2021autoformer}.}
    \vspace{-0.7cm}
    \label{tab:search_space}
    \small
    \begin{center}
    \setlength{\tabcolsep}{0.4em}
    \resizebox{\linewidth}{!}{%
      \begin{tabular}{lccc}
          \toprule
           & AutoFormer-T & AutoFormer-S & AutoFormer-B \\
          \midrule
          Head Num & [3, 4] & [5, 6, 7] & [9, 10] \\
          MLP Ratio & [3.5, 4] & [3, 3.5, 4] & [3, 3.5, 4] \\
          Embedding Dim & [192, 216, 240] & [320, 384, 448] & [512, 576, 624] \\
          Depth & [12, 13, 14] & [12, 13, 14] & [14, 15, 16] \\
          \bottomrule
      \end{tabular}
    }
    \end{center}
    \vspace{-0.4cm}
  \end{table}
  
\subsection{Experimental settings}
\label{sec:settings}

\subsubsection{Search space and datasets}
We evaluate TAS-LoRA on three variants of AutoFormer search space~\cite{chen2021autoformer} with different scales: AutoFormer-T, AutoFormer-S, and AutoFormer-B. They differ in the number of attention heads, MLP ratio, embedding dimensions, and network depths~(Table~\ref{tab:search_space}). For training and evaluation, we mainly use the ImageNet dataset~\cite{deng2009imagenet}. We also fine-tune the searched subnets on CIFAR-10/100~\cite{krizhevsky2009learning}, Flowers~\cite{nilsback2008flowers}, CARS~\cite{krause2013cars}, and INAT-19~\cite{van2018inaturalist} to assess the transferability and generalization ability of TAS-LoRA.

\begin{table}[t]
    \centering
    \captionsetup{font={small}}
    \caption{Quantitative results of ViT models on ImageNet~\cite{deng2009imagenet}. For TAS-LoRA, we report the average over 3 runs. $\dagger$: Results reproduced with the official code.  $*$: Require retraining for changes in hardware constraints. }

    \vspace{-0.2cm}
    \label{table:main_results}
    \footnotesize
    \renewcommand{\arraystretch}{0.95}
    \setlength{\tabcolsep}{0.5em}
  
      \begin{tabular}{lccc}
        \toprule
        Models & Top-1 (\%) & \#Params & FLOPs \\
        \midrule
        ViT-Ti~\cite{dosovitskiy2020image} & 74.5 & 5.7M & - \\
        DeiT-Ti~\cite{touvron2021training} & 72.2 & 5.7M & 1.2G \\
        ConViT-Ti~\cite{d2021convit} & 73.1 & 6.0M & 1.0G \\
        TNT-Ti~\cite{han2021transformer} & 73.9 & 6.1M & 1.4G \\
        TF-TAS-Ti$^*$~\cite{zhou2022training} & 75.3 & 5.9M & 1.4G \\
        AutoProx$^*$~\cite{wei2024auto} & 75.6 & 6.4M & - \\
        GLiT-Ti$^*$~\cite{chen2021glit} & 76.3 & 7.2M & 1.4G \\
        ViTAS-C~\cite{su2022vitas} & 74.7 & 5.6M & 1.3G \\
        FocusFormer-Ti$^*$~\cite{liu2022focusformer} & 75.1 & 6.2M & 1.4G \\
        PreNAS-Ti$^*$~\cite{PreNAS} & 77.1 & 5.9M & 1.4G \\
        AutoFormer-Ti~\cite{chen2021autoformer} & 74.7 & 5.7M & 1.3G \\
        AutoFormer-Ti$^\dagger$ & 74.7 & 5.9M & 1.3G \\
        \rowcolor[HTML]{DAE8FC}\textbf{TAS-LoRA-Ti} & 75.7 $\pm$ 0.1 & 5.9M & 1.3G \\
        DYNAS~\cite{jeon2025subnet} & 74.8 & 6.0M & 1.3G \\
        \rowcolor[HTML]{DAE8FC}\textbf{TAS-LoRA + DYNAS} & 75.9 $\pm$ 0.0 & 5.9M & 1.3G \\
        
        \midrule
        ViT-S~\cite{dosovitskiy2020image} & 78.8 & 22.1M & 4.7G \\
        DeiT-S~\cite{touvron2021training} & 79.9 & 22.1M & 4.7G \\
        ConViT-S~\cite{d2021convit} & 81.3 & 27.0M & 5.4G \\
        TNT-S~\cite{han2021transformer} & 81.5 & 23.8M & 5.2G \\
        TF-TAS-S$^*$~\cite{zhou2022training} & 81.9 & 22.8M & 5.0G \\
        GLiT-S$^*$~\cite{chen2021glit} & 80.5 & 24.6M & 4.4G \\
        ViTAS-F~\cite{su2022vitas} & 80.5 & 27.6M & 6.0G \\
        FocusFormer-S$^*$~\cite{liu2022focusformer} & 81.6 & 23.7M & 5.0G \\
        PreNAS-S$^*$~\cite{PreNAS} & 81.8 & 22.9M & 5.1G \\
        AutoFormer-S~\cite{chen2021autoformer} & 81.7 & 22.9M & 5.1G \\
        AutoFormer-S$^\dagger$ & 81.6 & 22.9M & 5.1G \\
        \rowcolor[HTML]{DAE8FC}\textbf{TAS-LoRA-S} & 81.9 $\pm$ 0.1 & 22.9M & 5.0G \\

        \midrule
        PVT-Large~\cite{wang2021pyramid} & 81.7 & 61.0M & 9.8G \\
        DeiT-B~\cite{touvron2021training} & 81.8 & 86.0M & 18.0G \\
        ViT-B~\cite{dosovitskiy2020image} & 79.7 & 86.0M & 18.0G \\
        ConViT-B~\cite{d2021convit} & 82.4 & 86.0M & 17.0G \\
        TF-TAS-B$^*$~\cite{zhou2022training} & 82.2 & 54.0M & 12.0G \\
        GLiT-B$^*$~\cite{chen2021glit} & 82.3 & 96.0M & 17.0G \\
        FocusFormer-B$^*$~\cite{liu2022focusformer} & 81.9 & 52.7M & 11.0G \\
        PreNAS-B$^*$~\cite{PreNAS} & 82.6 & 54.0M & 11.0G \\
        AutoFormer-B~\cite{chen2021autoformer} & 82.4 & 54.0M & 11.0G \\
        AutoFormer-B$^\dagger$ & 82.4 & 54.0M & 11.0G \\
        \rowcolor[HTML]{DAE8FC}\textbf{TAS-LoRA-B} & 82.6 $\pm$ 0.0 & 54.0M & 11.0G \\
        \bottomrule
    \end{tabular}
    \vspace{-0.4cm}
  \end{table}

  \begin{table*}[t]
    \centering
    \small
    \captionsetup{font={small}}
    \caption{Quantitative results of transfer learning on various datasets, including CIFAR-10/100~\cite{krizhevsky2009learning}, Flowers~\cite{nilsback2008flowers}, CARS~\cite{krause2013cars}, and INAT-19~\cite{van2018inaturalist}. All images are resized to $224 \times 224$.}
    \vspace{-0.4cm}
    \setlength{\tabcolsep}{0.2em}
    \renewcommand{\arraystretch}{0.9}
    \resizebox{0.5\linewidth}{!}{%
    \begin{tabular}{l c c c c c c}
        \toprule
        Model & \#Params (M) & CIFAR-10 & CIFAR-100 & Flowers & Cars & INAT-19 \\
        \midrule
        ViT-B/16~\cite{dosovitskiy2020image} & 86M & 98.1 & 87.1 & 89.5 & - & - \\
        ViT-L/16~\cite{dosovitskiy2020image} & 307M & 97.9 & 86.4 & 89.7 & - & - \\
        DeiT-B~\cite{touvron2021training} & 86M & 99.1 & 90.8 & 98.4 & 92.1 & 77.7 \\
        ViTAE-S~\cite{xu2021vitae} & 24M & 98.8 & 90.8 & 97.8 & 91.4 & 76.0 \\
        DearKD-S~\cite{chen2022dearkd} & 22M & 98.4 & 89.3 & 97.4 & 91.3 & - \\
        PreNAS-S~\cite{PreNAS} & 23M & 99.1 & 91.2 & 97.6 & 92.2 & 76.4 \\
        AutoFormer-S~\cite{chen2021autoformer} & 23M & 98.9 & 89.6 & 97.9 & 92.1 & 77.4 \\
        \rowcolor[HTML]{DAE8FC}
        \textbf{TAS-LoRA-S}& 23M  & 99.1 & 91.0 & 98.2 & 92.3 & 78.0 \\
        \bottomrule
    \end{tabular}
    }
    \vspace{-0.3cm}
    
    \label{tab:transfer}
\end{table*}

\subsubsection{Implementation details}
\textbf{Training MoLE.}
We freeze the pretrained supernet weights from AutoFormer~\cite{chen2021autoformer}, and train the MoLE layers and the router for 50 epochs. During a warm-up phase of 5 epochs, we update LoRA experts only with the AdamW optimizer~\cite{loshchilov2017decoupled}, while keeping the router frozen, encouraging each expert to learn unique characteristics from the corresponding group. Following the standard warm-up strategies~\cite{touvron2021training,chen2021autoformer}, the learning rate for LoRA experts starts with 1e-5, and increases linearly to 5e-4. After the warm-up, we optimize both the router and LoRA experts jointly for the remaining 45 epochs. The router uses SGD with an initial learning rate of 1e-1. We apply a cosine learning rate scheduler, and use a batch size of 1024.

We set the rank of the LoRA $r$ to 8. The number of LoRA experts $K$ is set to that of groups, obtained with the criteria of the number of attention heads, MLP ratio, and embedding dimension from transformer blocks. For example, AutoFormer-T includes 2 candidate values for the number of attention heads, 2 for the MLP ratio, and 3 for the embedding dimension, resulting in $12$ groups. That is, we set $K = 12$ for AutoFormer-T, $K = 27$ for AutoFormer-S, and $K = 18$ for AutoFormer-B.


\noindent \textbf{Router.} 
We implement a lightweight LSTM-based router to compute subnet-specific expert weights. Specifically, we use a single-layer LSTM with a hidden dimension of 128, followed by a fully connected layer as the router head. For block embedding, we adopt an embedding layer~\cite{mikolov2013efficient, pennington2014glove} encoding discrete architectural attributes into 256-dimensional vectors. We set $\beta$ to 3 for group-wise router initialization.


\noindent \textbf{Transfer learning.}
We follow the protocol in~\cite{touvron2021training} for transfer learning. Specifically, we fine-tune the searched subnet, except for the classifier, which is trained from scratch. For CIFAR-10/100~\cite{krizhevsky2009learning}, Flowers~\cite{nilsback2008flowers}, and CARS~\cite{krause2013cars}, we fine-tune the network for 1000 epochs using SGD, with an initial learning rate of 1e-2 and a weight decay of 1e-4. For INAT-19~\cite{van2018inaturalist}, subnets are fine-tuned for 360 epochs using AdamW~\cite{loshchilov2017decoupled} with an initial learning rate of 7.5e-5 and a weight decay of 5e-2.

\subsection{Results}
\textbf{ImageNet.}
We compare in Table~\ref{table:main_results} TAS-LoRA with the state-of-the-art ViT models, in terms of the top-1 accuracy, the number of parameters, and FLOPs. We summarize our findings as follows: (1) TAS-LoRA consistently improves the performance of supernets trained with AutoFormer~\cite{chen2021autoformer} across all search spaces. This suggests that alleviating the feature collapse problem by learning subnet-specific features with TAS-LoRA can enhance the performance. (2) TAS-LoRA enhances the performance of DYNAS~\cite{jeon2025subnet}, which also exploits a weight-entangled supernet with a different training strategy. This indicates that our method is effective in addressing the feature collapse problem, regardless of the training strategy. Note that TAS-LoRA can be applied to supernets trained with other TAS methods, which is however not feasible, since official implementations of these methods are either unavailable~\cite{liu2022focusformer} or incomplete~\cite{chen2021glit,su2022vitas}\footnote{The key components of these approaches not fully disclosed in the official repositories, and related issues are raised in the corresponding issue trackers.}. (3) The performance gain of TAS-LoRA is more significant in smaller spaces~(\eg,~AutoFormer-T). Note that a supernet is constructed to cover all possible architecture configurations within a search space, and the number of parameters is determined by the largest subnet supported by the search space. The largest subnet in AutoFormer-T is smaller, compared to those in larger spaces~(\eg,~AutoFormer-S/B), suggesting that the number of parameters for the supernet in AutoFormer-T is relatively small. This reduces the representational capacity of the supernet, and limits representing diverse features across subnets, worsening the feature collapse problem. TAS-LoRA can effectively mitigate the problem with subnet-specific feature learning, which leads to notable improvements. (4) TAS-LoRA achieves a better trade-off between complexity and accuracy, compared to other methods. For example, the GLiT-B has 96.0M parameters with a top-1 accuracy of 82.3\%, while TAS-LoRA-B achieves a higher accuracy of 82.6\% with only 54M parameters. This demonstrates the effectiveness of TAS-LoRA in learning subnet-specific features, leading to better overall performance. (5) TAS-LoRA underperforms PreNAS~\cite{PreNAS} in the AutoFormer-T space. PreNAS constructs a supernet with nearly the same number of parameters as the original supernet, but reduces the number of subnets significantly~(\eg,~6 out of $2\times10^8$ for AutoFormer-T). Accordingly, the shared weights in the supernet would be optimized better for the few subnets, partially alleviating the feature collapse problem. However, PreNAS is less flexible under changes in hardware constraints. Specifically, if hardware constraints~(\eg,~latency and FLOPs) are altered, and the initially selected subnets do not satisfy them, PreNAS should retrain the entire supernet from scratch. In contrast, TAS-LoRA trains a supernet including all possible subnets with subnet-specific feature learning, enabling adapting to various constraints without retraining.

\noindent \textbf{Transfer learning.}
We show in Table~\ref{tab:transfer} the transfer learning results of TAS-LoRA on various datasets, including CIFAR-10/100~\cite{krizhevsky2009learning}, Flowers~\cite{nilsback2008flowers}, CARS~\cite{krause2013cars}, and INAT-19~\cite{van2018inaturalist}. TAS-LoRA achieves competitive or superior performance on all datasets, which cover a wide range of domains such as natural objects~\cite{krizhevsky2009learning}, fine-grained categories~\cite{nilsback2008flowers, krause2013cars}, and large-scale species classification~\cite{van2018inaturalist}, demonstrating the generalization ability of TAS-LoRA.

\begin{table}[t]
    \captionsetup{font={small}}
    \caption{Quantitative comparison of different router initialization methods for the AutoFormer search space~\cite{chen2021autoformer} on ImageNet~\cite{deng2009imagenet}, in terms of the top-1 validation accuracy (\%).}
    \vspace{-0.5cm}
    \label{tab:group_wise_init_ablations}
    \small
    \vspace{-1.8mm}
    \begin{center}
    \setlength{\tabcolsep}{0.2em}
    \resizebox{0.7\linewidth}{!}{%
    \begin{tabular}{lcccc}
      \toprule
      Methods & Router Init. & Tiny & Small & Base \\
      \midrule
      AutoFormer~\cite{chen2021autoformer} & - & 74.9 & 81.6 & 82.4 \\
      TAS-LoRA & Random & 75.4 & 81.8 & 82.5 \\
      TAS-LoRA & Group-wise & 75.7 & 81.9 & 82.6 \\ 
      \bottomrule
    \end{tabular}
    }
    \end{center}
    \vspace{-0.5cm}
  \end{table}

\subsection{Analysis}

\begin{figure}[t]
    \centering
    \captionsetup{font={small}}
    \includegraphics[width=0.55\linewidth]{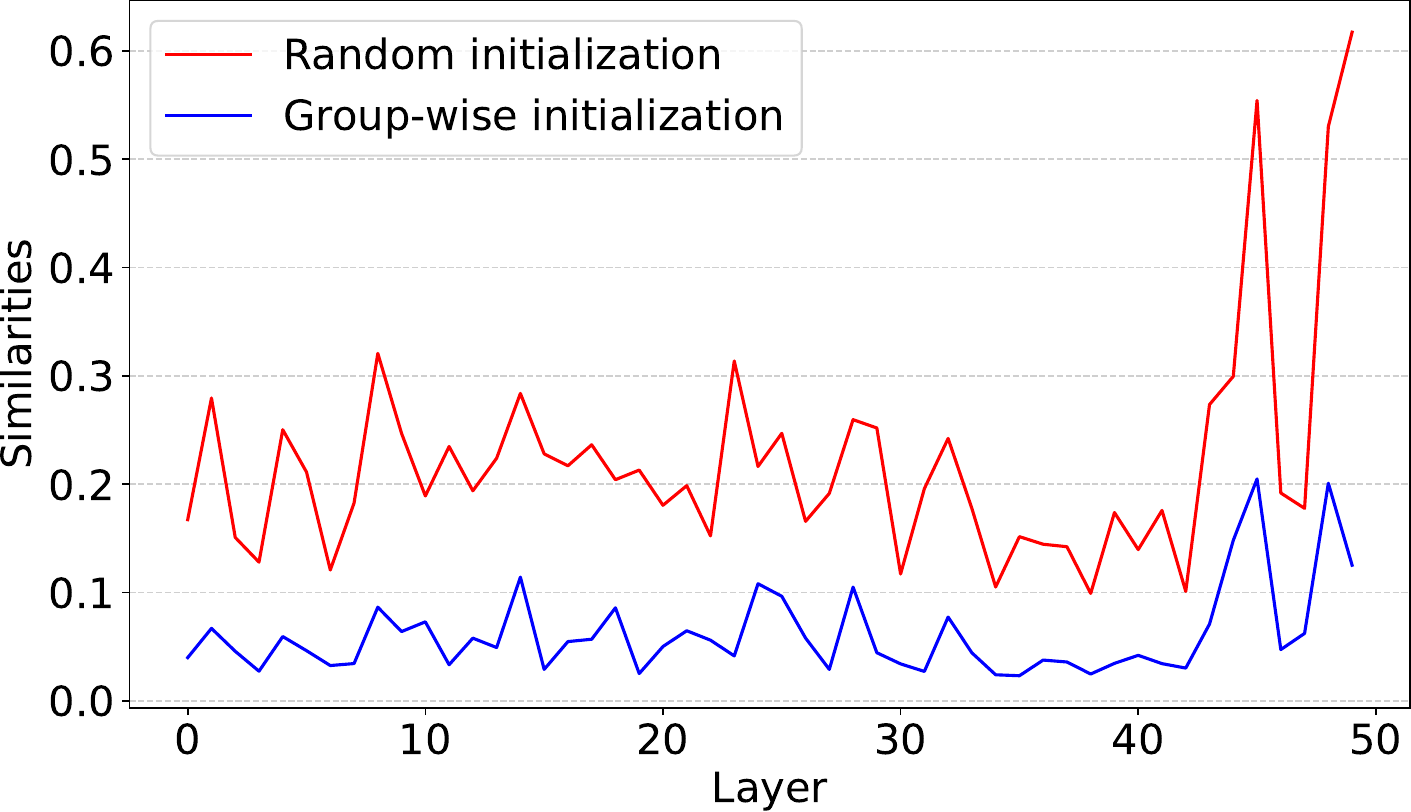}
    \vspace{-0.4cm}
    \caption{Cosine similarities of features from LoRA experts across layers. We use supernets for the AutoFormer-T space~\cite{chen2021autoformer} on ImageNet~\cite{deng2009imagenet}, trained using TAS-LoRA with random and group-wise router initialization. For each layer, we calculate the average cosine similarity across all expert combinations.}
    \label{fig:complexity-wise-sim}
    \vspace{-0.5cm}
\end{figure}

\begin{figure}[t]
    \centering
    \captionsetup{font={small}}

    \begin{subfigure}[t]{0.42\linewidth}
        \centering
        \includegraphics[width=\linewidth]{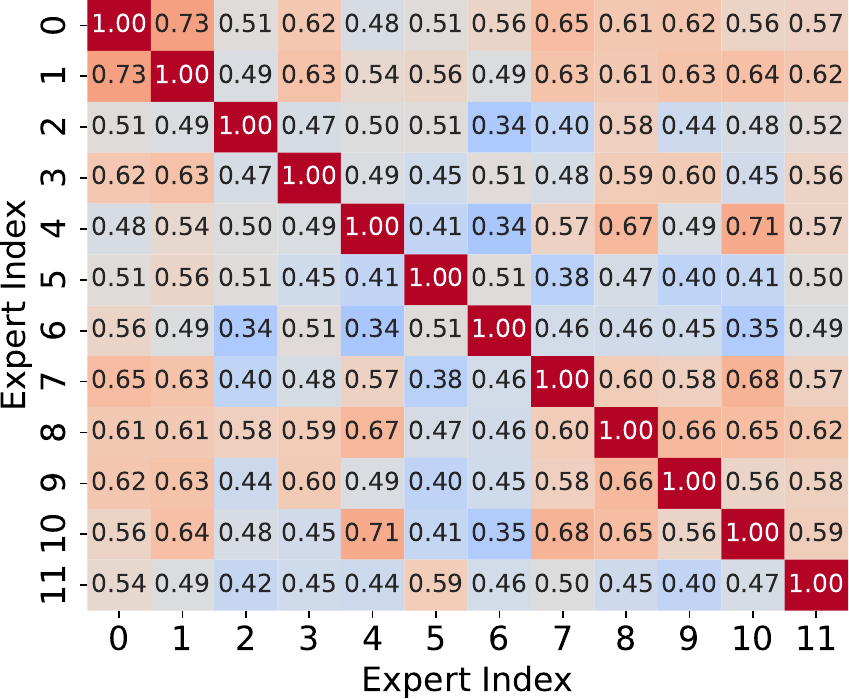}
        \vspace{-0.35cm}
        \caption{Random init.}
    \end{subfigure}
    \hfill
    \begin{subfigure}[t]{0.42\linewidth}
        \centering
        \includegraphics[width=\linewidth]{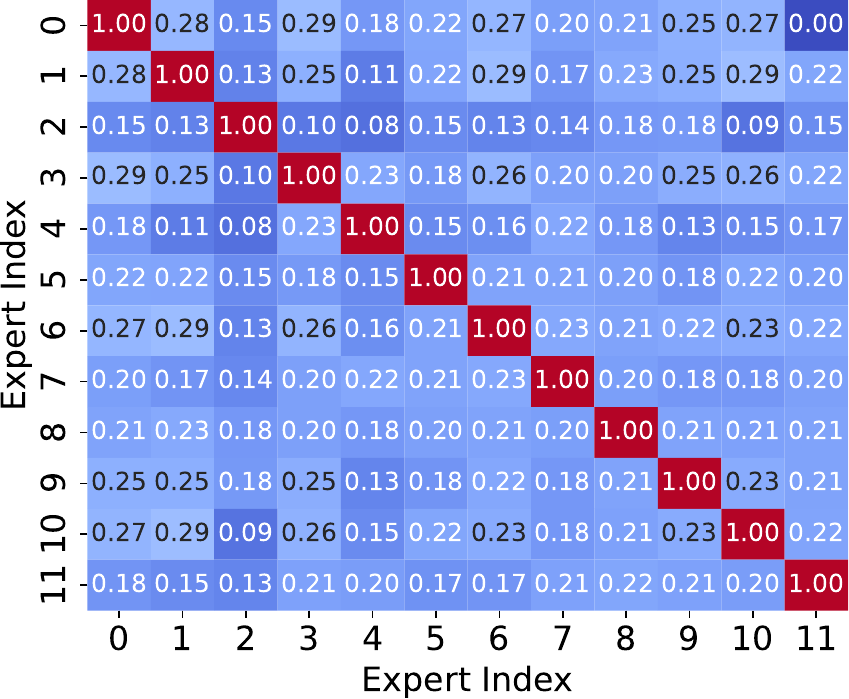}
        \vspace{-0.35cm}
        \caption{Group-wise init.}
    \end{subfigure}
    \hfill
    \begin{subfigure}[t]{0.045\linewidth}
        \centering
        \vspace{-2.9cm}
        \includegraphics[width=\linewidth]{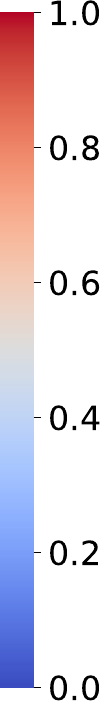}
    \end{subfigure}

    \vspace{-0.35cm}
    \caption{Cosine similarities of features from LoRA experts in the QKV projection layer of the $12^{\text{th}}$ block. We use supernets for the AutoFormer-T space~\cite{chen2021autoformer} on ImageNet~\cite{deng2009imagenet}, trained using TAS-LoRA with random and group-wise initialization methods.}
    \label{fig:sim}
    \vspace{-0.2cm}
\end{figure}

\noindent \textbf{Group-wise router initialization.}
We compare in Table~\ref{tab:group_wise_init_ablations} the performance for router initialization methods. We can see that TAS-LoRA with a group-wise initialization technique outperforms the random one across all search spaces. This suggests that reducing the redundancy for experts gives better performance, providing diverse feature representations effectively. To validate this, we measure cosine similarities of features, provided by LoRA experts, across layers in Fig.~\ref{fig:complexity-wise-sim}. It shows that the group-wise initialization gives lower similarities between experts, compared to the random one, confirming that our method encourages learning diverse feature representations across experts. We also visualize in Fig.~\ref{fig:sim} cosine similarities of features from LoRA experts in the QKV projection layer of the 12$^{th}$ block. With the random initialization, LoRA experts give highly similar features, while the group-wise one leads to more distinct representations.

\begin{figure}[t]
    \centering
    \captionsetup{font={small}}
    \includegraphics[width=0.7\linewidth]{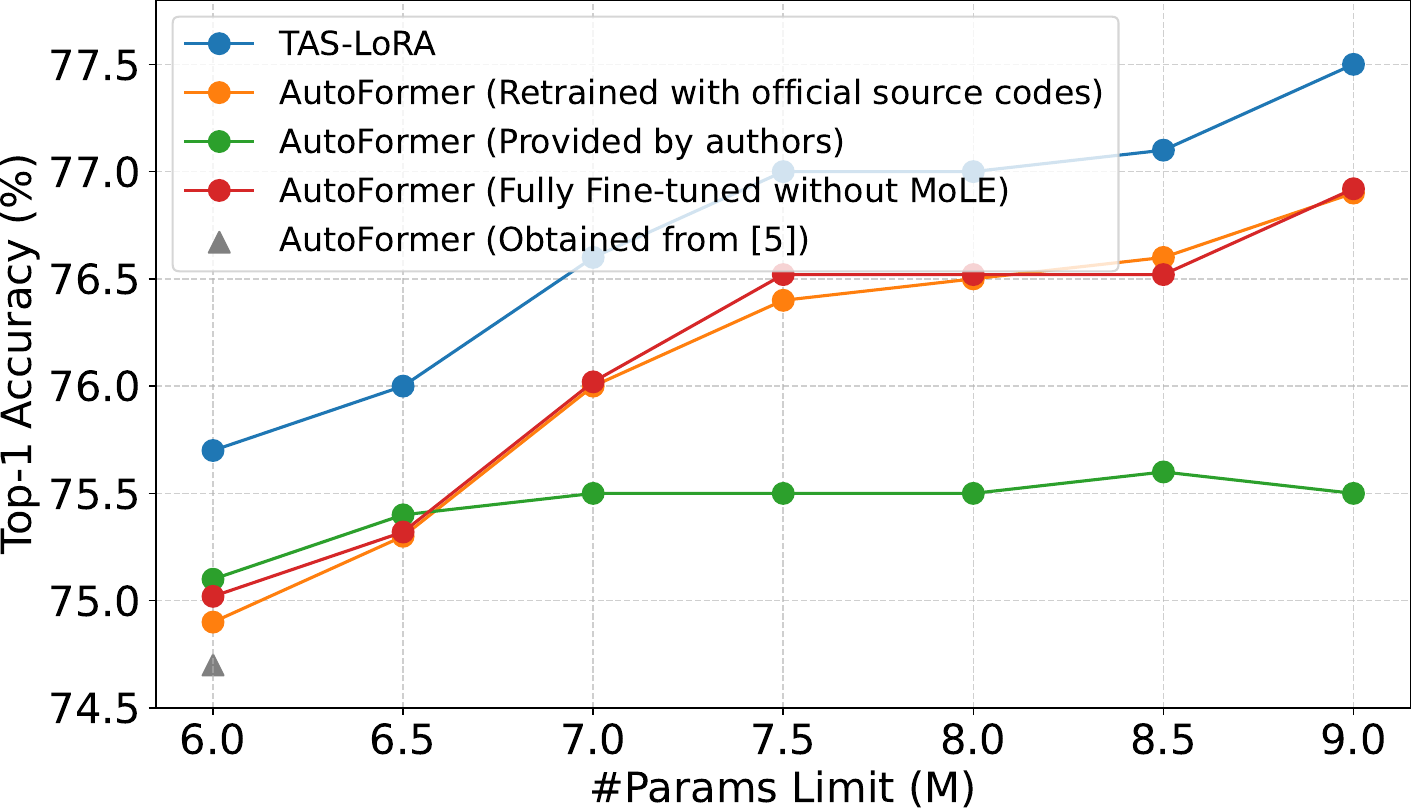}
    \vspace{-0.3cm}
    \caption{Top-1 accuracies on ImageNet~\cite{deng2009imagenet} in the AutoFormer-T space~\cite{chen2021autoformer} under various parameter limits. We compare TAS-LoRA with AutoFormer~\cite{chen2021autoformer}: (1) a retrained model with the official code; (2) a supernet provided by the authors; (3) a full fine-tuned version for the retrained supernet; (4) the result reported in AutoFormer.}
    \label{fig:complexity-wise-acc}
    \vspace{-0.4cm}
  \end{figure}

  \vspace{-0.1cm}
  \noindent \textbf{Comparison with fine-tuning.}
TAS-LoRA trains MoLE layers and the router only for 50 epochs, while keeping original weights in the supernet frozen. To validate its effectiveness, we compare TAS-LoRA with a full fine-tuned model, where all supernet weights are updated for the same number of epochs without adopting MoLE. We show in Fig.~\ref{fig:complexity-wise-acc} top-1 accuracies of searched subnets across various parameter limits. We observe that TAS-LoRA consistently outperforms the fully fine-tuned model across various parameter limits, demonstrating that the performance gains come from learning subnet-specific features through MoLE, rather than simply extending training epochs. In contrast, the full fine-tuning model shows almost no improvement over the pretrained supernet, indicating that it is ineffective in addressing the feature collapse problem. We can also see that the supernet provided by the work of~\cite{chen2021autoformer} performs well under small parameter limits, but shows notable performance degradation for larger ones, suggesting that the supernet might be optimized for small ones. To avoid this issue, we retrain the supernet using the official source code provided by the authors before applying TAS-LoRA.

\section{Conclusion}
We have identified the feature collapse problem in TAS, where subnets fail to learn distinct feature representations due to the shared weights in a supernet. We have proposed TAS-LoRA, a novel method that leverages LoRA to enable subnet-specific feature learning. By exploiting LoRA experts with a dynamic routing mechanism, TAS-LoRA enables subnets within a supernet to learn specialized feature representations efficiently. Furthermore, we present a group-wise router initialization technique to reduce the redundancy in experts, and promote diverse expert training from the early stages.  We have shown extensive experimental results to validate the effectiveness of TAS-LoRA. 

\section*{Acknowledgements}
 This work was partly supported by IITP grant funded by the Korea government (MSIT) (No.RS2022-00143524, Development of Fundamental Technology and Integrated Solution for NextGeneration Automatic Artificial Intelligence System, No.2022-0-00124, RS-2022-II220124, Development of Artificial Intelligence Technology for Self-Improving Competency-Aware Learning Capabilities, AI Semiconductor Innovation Lab (Yonsei University)), and the KIST Institutional Program (Project No.2E33001-24-086).

{
    \small
    \bibliographystyle{ieeenat_fullname}
    \bibliography{main}
}
\newpage

\renewcommand{\theequation}{\roman{equation}} 
\setcounter{equation}{0} 

\renewcommand{\thefigure}{\Alph{figure}} 
\setcounter{figure}{0} 

\renewcommand{\thetable}{\Alph{table}} 
\setcounter{table}{0} 

\renewcommand{\thesection}{\Alph{section}} 
\setcounter{section}{0} 

\maketitlesupplementary

In this supplementary material, we provide more detailed analyses on the design of TAS-LoRA~(Sec.~\ref{sec:design_analysis}), additional experiments for searching hyperparameters~(Sec.~\ref{sec:more_experiments}), and limitations of our work and future directions~(Sec.~\ref{sec:limitations}). We then summarize the training, and search procedures of TAS-LoRA~(Sec.~\ref{sec:algorithm}). 
Finally, we provide the computational resources used in our experiments~(Sec.~\ref{resources}).

\section{Analyses on design of TAS-LoRA}
\label{sec:design_analysis}
In this section, we provide a more detailed analysis of the design choices in TAS-LoRA.

\subsection{Router architecture}
We show in Table~\ref{tab:router_design} the results of different router architectures in the AutoFormer-T search space. In all cases, the router takes as input the sequence of architectural attributes of the sampled subnet~(\eg, MLP ratio, number of attention heads, and embedding dimension). The router then predicts routing weights for the MoLE experts for each block. We compare recurrent routers, which can capture sequential dependencies across blocks, with a non-recurrent MLP baseline that treats each block independently. We can see that recurrent routers, \ie, LSTM~\cite{hochreiter1997long} and GRU~\cite{chung2014empirical}, perform similarly, while the block-wise MLP performs worse. This indicates that modeling dependencies across transformer blocks is beneficial for expert routing.

\subsection{MoLE}
We show in Table~\ref{tab:mole_vs_singlelora} the comparison between using a single shared LoRA module for all subnets and the proposed MoLE.
For a fair comparison, both variants are inserted into the same target layers, and are trained under the same search and optimization setting. We can see that replacing MoLE with a single shared LoRA degrades the performance from 75.7 to 75.0. This is because, since a single LoRA is shared across all subnets, which still results in the feature collapse problem. Our MoLE mitigates this issue by maintaining multiple experts and using the router to assign different experts to different subnets based on their architectural attributes, allowing for more specialized adaptation.

\begin{table}[t]
    \centering
    \small
    \caption{Comparison of different router architectures in the AutoFormer-T search space. All routers use the same routing input and are trained with the same optimization setting.}
    \label{tab:router_design}
    \begin{tabular}{lc}
        \toprule
        Router & Top-1 (\%) \\
        \midrule
        LSTM & 75.7 \\
        GRU  & 75.6 \\
        Block-wise MLP  & 75.4 \\
        \bottomrule
    \end{tabular}
\end{table}
\begin{table}[t]
    \centering
    \small
    \caption{Comparison between a single shared LoRA and the proposed mixture-of-LoRA experts in the AutoFormer-T search space.}
    \label{tab:mole_vs_singlelora}
    \begin{tabular}{lc}
        \toprule
        Method & Top-1 (\%) \\
        \midrule
        Single LoRA & 75.0 \\
        TAS-LoRA (MoLE) & 75.7 \\
        \bottomrule
    \end{tabular}
\end{table}

\subsection{Group-wise initialization strategy}
We show in Table~\ref{tab:grouping_strategy} the results of different grouping strategies used in router initialization. In TAS-LoRA, we group blocks by architecture similarities, and assign each group to a different expert during the early stage of training. We compare this strategy with random grouping and grouping by the number of parameters. We can see that grouping by architectural similarity achieves the best performance. This is because grouping by architectural similarity reduces redundancy among experts and encourages diverse feature learning across experts early in training.

\begin{table}[t]
    \centering
    \small
    \caption{Comparison of different grouping strategies for router initialization in the AutoFormer-T search space.}
    \label{tab:grouping_strategy}
    \begin{tabular}{lc}
        \toprule
        Grouping strategy & Top-1 (\%) \\
        \midrule
        Architectural similarity & 75.7 \\
        Random grouping & 75.5 \\
        Number of parameters & 75.6 \\
        \bottomrule
    \end{tabular}
\end{table}

\subsection{Architectural attributes for routing}
We show in Table~\ref{tab:routing_attributes} the ablation results for choosing the architectural attributes used by the router. In TAS-LoRA, the router predicts expert weights for each block based on the architectural properties of the sampled subnet. To this end, we provide the router with block-level and subnet-level attributes, including the MLP ratio, the number of attention heads, the embedding dimension, and the depth index. Using a richer set of attributes allows the router to distinguish candidate subnets more precisely, but it also increases the number of distinct groups considered in the initialization stage and may require a larger number of experts.

In this ablation, we incrementally vary the set of routing attributes while keeping the rest of the framework unchanged. We first consider the full set of attributes, and then remove some of them to study whether a more compact routing input can still preserve the effectiveness of subnet-specific expert assignment.

We can see that using richer architectural attributes generally improves performance, while using only a limited subset leads to degraded results. In particular, using only the number of heads, or the number of heads together with the embedding dimension, does not provide sufficient information for the router to distinguish subnets effectively. We also observe that adding depth gives only a marginal gain over using the MLP ratio, number of heads, and embedding dimension. Since including depth increases the number of distinct groups in the initialization stage and thus enlarges the expert space, we exclude depth in the main experiments and use the remaining attributes as the default routing input.

\begin{table}[t]
    \centering
    \small
    \caption{Ablation on the architectural attributes used for routing in the AutoFormer-T search space.}
    \label{tab:routing_attributes}
    \resizebox{0.95\columnwidth}{!}{
    \begin{tabular}{cccc|c}
        \toprule
        MLP ratio & \#Head & Embed. dim. & Depth & Top-1 (\%) \\
        \midrule
        \checkmark & \checkmark & \checkmark & \checkmark & 75.8 \\
        \checkmark & \checkmark & \checkmark &            & 75.7 \\
                   & \checkmark & \checkmark &            & 75.5 \\
                   & \checkmark &            &            & 75.2 \\
        \bottomrule
    \end{tabular}
    }
\end{table}

\section{Hyperparameters}
\label{sec:more_experiments}
In this section, we investigate the effect of several hyperparameters: (1) LoRA rank, (2) the $\beta$ value in router initialization, (3) learning rates for MoLE and the router, and (4) the number of warm-up epochs. Notably, TAS-LoRA consistently outperforms AutoFormer~\cite{chen2021autoformer} across all settings, demonstrating its effectiveness.

\paragraph{LoRA rank.}
We present in Table~\ref{tab:rank_results} the results of TAS-LoRA with varying LoRA rank $r$. We observe that TAS-LoRA achieves the best performance when the rank is set to 8 across all search spaces. We attribute this to the following: (1) a small rank may not provide sufficient capacity for capturing subnet-specific features, while (2) a large rank increases the number of trainable parameters, which can lead to overfitting. We also compare in Table~\ref{tab:rank_results} the number of trainable parameters of TAS-LoRA to that of full fine-tuning. We can see that when $r$ is set to 2, 4, or 8, TAS-LoRA requires fewer parameters than full fine-tuning, while with $r=16$, the parameter count exceeds that of full fine-tuning. Considering this trade-off between performance and parameter efficiency, we set the rank to 8 in all experiments.

\paragraph{$\beta$ in router initialization.}
We present in Table~\ref{tab:beta_results} the results of TAS-LoRA with different $\beta$ values in router initialization. We can see that TAS-LoRA achieves the best performance when $\beta$ is set to 3. If $\beta$ is too small, the router may fail to promote sufficient specialization among experts. Conversely, if $\beta$ is too large, it may reduce the flexibility of expert selection, limiting the diversity of experts assigned to subnets. Based on this, we set $\beta=3$ in all experiments.

\paragraph{Learning rates}
We show in Table~\ref{tab:lr_results} the results of TAS-LoRA with different learning rates for the MoLE and router. We observe that the best performance is obtained when the learning rate is set to 5e-4 for the MoLE and 1e-1 for the router, and use them in all experiments.

\paragraph{Warm-up epochs}
We present in Table~\ref{tab:warmup_results} the results of TAS-LoRA with different numbers of warm-up epochs. TAS-LoRA achieves the best performance when the number of warm-up epochs is set to 5. A shorter warm-up limits the time for the experts to specialize in their assigned groups, while a longer warm-up may hinder the joint optimization of the router and MoLE. Based on the results, we set the number of warm-up epochs to 5 in all experiments.

    \begin{table}[t]
        \centering
                \caption{Top-1 accuracy and parameter ratio across LoRA ranks in AutoFormer-T/S/B on ImageNet~\cite{deng2009imagenet}.}
        \resizebox{\linewidth}{!}{
        \begin{tabular}{c|cc|cc|cc}
            \toprule
            \multirow{2}{*}{\raisebox{-0.8ex}{Rank}} 
            & \multicolumn{2}{c|}{AutoFormer-T} & \multicolumn{2}{c|}{AutoFormer-S} & \multicolumn{2}{c}{AutoFormer-B} \\
            \cmidrule(lr){2-3} \cmidrule(lr){4-5} \cmidrule(lr){6-7}
            & Acc. (\%) & Params. (\%) & Acc. (\%) & Params. (\%) & Acc. (\%) & Params. (\%) \\
            \midrule
            2 & 75.5 & 23 & 81.7 & 32 & 82.4 & 11 \\
            4 & 75.6 & 35 & 81.8 & 48 & 82.5 & 18 \\
            8 & 75.7 & 59 & 81.9 & 79 & 82.6 & 33 \\
            16 & 75.6 & 107 & 81.7 & 142 & 82.5 & 63 \\
            \bottomrule
        \end{tabular}
        }

        \label{tab:rank_results}
    \end{table}

    \begin{table}[t]
        \centering
                \caption{Top-1 accuracy of TAS-LoRA in AutoFormer-T with varying $\beta$ values for router initialization.}
        \begin{tabular}{ccccc}
            \toprule
            $\beta$ & 1 & 3 & 10 \\
            \midrule
            Top-1 acc. (\%) & 75.5 & 75.7 & 75.2 \\
            \bottomrule
        \end{tabular}

        \label{tab:beta_results}
    \end{table}

    \vspace{0.5cm}

    \begin{table}[t]
        \centering
                \caption{Top-1 accuracy across learning rate combinations for MoLE ($\eta_L$) and router ($\eta_R$) in AutoFormer-T.}
        \renewcommand{\arraystretch}{1.2}
        \begin{tabular}{c||ccc}
            \toprule
            \diagbox{$\eta_L$}{$\eta_R$} & 2e-1 & 1e-1 & 5e-2 \\
            \midrule
            \midrule
            1e-3 & 75.7 & 75.6 & 75.7 \\
            5e-4 & 75.5 & 75.7 & 75.5 \\
            1e-4 & 75.3 & 75.3 & 75.2 \\
            \bottomrule
        \end{tabular}

        \label{tab:lr_results}
    \end{table}
    \begin{table}[t]
        \setlength{\tabcolsep}{1.5pt}

        \centering
               \caption{Top-1 accuracy of TAS-LoRA in AutoFormer-T with varying numbers of warm-up epochs.}
        \renewcommand{\arraystretch}{1.2}
        \begin{tabular}{cccccc}
            \toprule
            Warm-up epochs & 0 & 3 & 5 & 10 \\
            \midrule
            Top-1 acc. (\%) & 75.5 & 75.6 & 75.7 & 75.6 \\
            \bottomrule
        \end{tabular}
 
        \label{tab:warmup_results}
    \end{table}

\section{Limitations and future work}
\label{sec:limitations}
The number of LoRA experts in TAS-LoRA is directly tied to the number of groups, which corresponds to the number of candidate values for each architectural attribute in the search space. For example, if the number of attention heads varies across 4 candidate values, this results in 4 groups for that attribute. In large-scale search spaces with many attributes and extensive candidate sets, the total number of groups, and thus the number of experts can increase significantly, which may raise computational overhead. To address this, we plan to explore clustering techniques that merge similar candidates within each attribute (\eg, clustering attention head sizes with comparable capacities), thereby reducing the number of groups while ensuring subnet-specific feature learning.

In the current implementation, the number of experts and the rank of the LoRA modules are uniformly set across all layers. Recent studies on LoRA~\cite{zhang2023adalora,gao2024higher} suggest that adaptively assigning ranks and expert counts based on the importance of each layer can further enhance performance. As such, future work will explore dynamic expert allocation and rank adaptation tailored to individual layers to improve both efficiency and accuracy.

    \section{Algorithms}
    \label{sec:algorithm}
    We summarize in Algorithms~\ref{alg:lora_tas_training} and~\ref{alg:evolutionary_transformer_search} the training and search procedures of TAS-LoRA, respectively. For searching procedure, we use the evolutionary search algorithm, following the same procedure as in~\cite{chen2021autoformer}. Specifically, we sample 10,000 training examples as the validation set. We set the population size $P$ to 50, and number of iterations $T$ to 20. For each iteration, we pick top-10 subnets based on the validation loss, and generate the child networks by mutating and crossing over the top-10 subnets, with probabilities of 0.2, and 0.4, respectively. 

    \section{Computational resources}
\label{resources}
We conduct all experiments on NVIDIA A5000 and A6000 GPUs. For the AutoFormer-T space, we use 4 A5000 GPUs, while for AutoFormer-S and AutoFormer-B, we utilize 8 A5000 and 8 A6000 GPUs, respectively. During the searching process, we employ a single A5000 GPU. For transfer learning experiments, we use 4 A5000 GPUs.

    \begin{algorithm*}[h]
    \caption{Training of TAS-LoRA}
    \label{alg:lora_tas_training}
    \begin{algorithmic}[1]
    \STATE \textbf{Input}: Pretrained supernet weights $W$, LoRA experts $E$, Router $U$, training dataset $D_{\text{train}}$, max iteration $T$, number of transformer blocks $B$, number of layers $L$
    
    \STATE Initialize LoRA experts $E$ and router $U$
    \STATE Load pretrained supernet weights $W$
    
    \FOR{$t = 1$ to $T$}
        \STATE Sample a mini-batch from $D_\text{train}$ and a subnet architecture $\mathcal{N}_i$ from the supernet

\STATE Extract subnet-level attributes $(e, v)$ and block-level attributes $\{(n^{(b)}, m^{(b)})\}_{b=1}^{v}$

\STATE Compute expert weights $P$ with router $U$

\FOR{each layer $l = 1$ to $L$}
    \STATE Merge weights: $W^l_{\text{merged}} = W^l + \sum_{k=1}^{K} p^l_k E^l_k$
    
    \STATE Compute output: $y^l = W^l_{\text{merged}} x^l$
\ENDFOR

\STATE Compute loss $\mathcal{L}_{\text{train}}$

\STATE Update $E$ and $U$ using gradient descent on $\mathcal{L}_{\text{train}}$

    \ENDFOR
    \STATE \textbf{Return:} Trained LoRA experts and router
    \end{algorithmic}
    \end{algorithm*}

        \begin{algorithm*}[h] \caption{Searching of TAS-LoRA} \label{alg:evolutionary_transformer_search} \begin{algorithmic}[1] \STATE \textbf{Input}: Pretrained supernet weights $\theta$, LoRA and router parameters $\phi$, population size $P$, architecture constraints $C$, number of iterations $T$, validation dataset $D_{\text{val}}$, search space $\mathcal{N}$ 
            \STATE Initialize population $P_0$ with $P$ random subnets satisfying $C$ 
            
            \FOR{$i = 1$ to $T$}  
            \STATE $\mathcal{L}_{\text{val}} \gets \emptyset$
        \FOR{each subnet $\mathcal{N}_s$ in population $P_{i-1}$}
        \IF{$\mathcal{N}_s$ satisfies constraints $C$}
            \STATE Evaluate validation loss $\mathcal{L}_{\text{val}}(\mathcal{N}_s; \theta, \phi)$ on $D_{\text{val}}$
            \STATE Store $\mathcal{L}_{\text{val}}[\mathcal{N}_s] = \mathcal{L}_{\text{val}}(\mathcal{N}_s; \theta, \phi)$
        \ENDIF
    \ENDFOR

    \STATE Select top-$k$ subnets: $Top_k = \text{Select\_Topk}(P_{i-1}, \mathcal{L}_{\text{val}}, k)$

    \STATE Generate mutated subnets: $P_{\text{mutation}} = \text{Mutation}(Top_k, C)$
    
    \STATE Generate crossover subnets: $P_{\text{crossover}} = \text{Crossover}(Top_k, C)$
    
    \STATE Create next population $P_i = P_{\text{mutation}} \cup P_{\text{crossover}}$
    
    \STATE Fill remaining slots with random subnets: $P_i = P_i \cup \text{Random\_Candidates}(P - |P_i|, \mathcal{N}, C)$

\ENDFOR 

\STATE $\mathcal{N^*} =$ Select\_Topk$(P_T, \mathcal{L}_\text{val}, 1)$

\STATE \textbf{Return:} Optimal subnet $\mathcal{N}^*$

 \end{algorithmic} \end{algorithm*}




\end{document}